\documentclass[10pt,twocolumn,letterpaper]{article}

\usepackage[pagenumbers]{cvpr} 

\usepackage{graphicx}
\usepackage{amsmath}
\usepackage{amssymb}
\usepackage{booktabs}
\usepackage{bm}
\usepackage{enumerate}
\usepackage{caption}
\usepackage[accsupp]{axessibility}  

\usepackage[pagebackref,breaklinks,colorlinks]{hyperref}

\usepackage[capitalize]{cleveref}
\crefname{section}{Sec.}{Secs.}
\Crefname{section}{Section}{Sections}
\Crefname{table}{Table}{Tables}
\crefname{table}{Tab.}{Tabs.}

\def\cvprPaperID{1038} 
\def\confName{CVPR}
\def\confYear{2023}

\begin{document}

\title{Diffusion-SDF: Text-to-Shape via Voxelized Diffusion}

\author{Muheng Li$^1$, Yueqi Duan$^{\dag,2}$, Jie Zhou$^1$, Jiwen Lu$^1$\\
$^1$Department of Automation, Tsinghua University\\
$^2$Department of Electronic Engineering, Tsinghua University}

\twocolumn[{%
\renewcommand\twocolumn[1][]{#1}%
\maketitle
\begin{center}
    \centering
    \vspace{-15pt}
    \captionsetup{type=figure}
    \includegraphics[width = 1\linewidth]{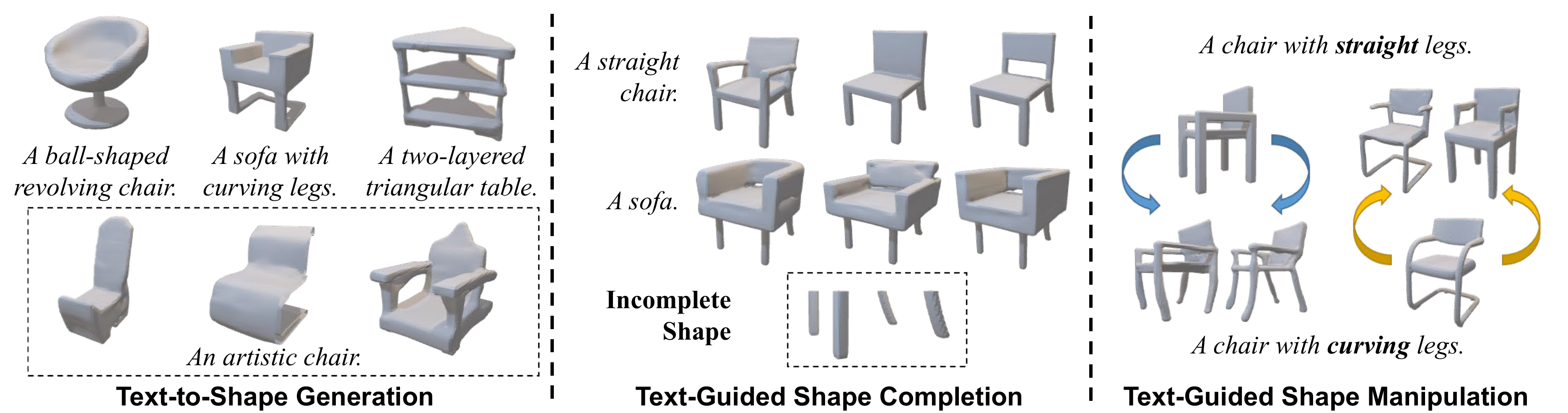}
    \captionof{figure}{We propose a text-to-shape synthesis approach named Diffusion-SDF. Our method is capable of performing various text-to-shape tasks including text-to-shape generation (directly generating 3D shapes from texts), text-guided shape completion (generating the missing part of an incomplete shape under the guidance of texts), and text-guided shape manipulation (modifying a given shape according to texts).}
    \label{fig:f0}
\end{center}%
}]



\begin{abstract}

With the rising industrial attention to 3D virtual modeling technology, generating novel 3D content based on specified conditions (\eg text) has become a hot issue. In this paper, we propose a new generative 3D modeling framework called Diffusion-SDF for the challenging task of text-to-shape synthesis. Previous approaches lack flexibility in both 3D data representation and shape generation, thereby failing to generate highly diversified 3D shapes conforming to the given text descriptions. To address this, we propose a SDF autoencoder together with the Voxelized Diffusion model to learn and generate representations for voxelized signed distance fields (SDFs) of 3D shapes. Specifically, we design a novel UinU-Net architecture that implants a local-focused inner network inside the standard U-Net architecture, which enables better reconstruction of patch-independent SDF representations. We extend our approach to further text-to-shape tasks including text-conditioned shape completion and manipulation. Experimental results show that Diffusion-SDF generates both higher quality and more diversified 3D shapes that conform well to given text descriptions when compared to previous approaches. Code is available at: \url{https://github.com/ttlmh/Diffusion-SDF}.
\end{abstract}
\let\thefootnote\relax\footnotetext{$^{\dag}$Corresponding author. }
\section{Introduction}
\label{sec:intro}
Exploring data representations for 3D shapes has been a fundamental and critical issue in 3D computer vision. Explicit 3D representations including point clouds~\cite{QiSMG17, QiYSG17}, polygon meshes~\cite{GroueixFKRA18, KatoUH18} and occupancy voxel grids~\cite{WuSKYZTX15, ChoyXGCS16} have been widely applied in various 3D downstream applications~\cite{AchlioptasDMG18, YangHH0BH19, NashGEB20}. While explicit 3D representations achieve encouraging performance, there are some primary limitations including not being suitable for generating watertight surfaces (\eg point clouds), or being subject to topological constraints (\eg meshes). On the other hand, implicit 3D representations have been widely studied more recently~\cite{ParkFSNL19, ChabraLISSLN20, DuanZ0YNG20}, with representative works including DeepSDF~\cite{ParkFSNL19}, Occupancy Network~\cite{MeschederONNG19} and IM-Net~\cite{ChenZ19}. In general, implicit functions encode the shapes by the iso-surface of the function, which is a continuous field and can be evaluated at arbitrary resolution.
In recent years, numerous explorations have been conducted for implicit 3D generative models, which show promising performance on several downstream applications such as single/multi-view 3D reconstruction~\cite{0002JHZ20, XuWCMN19} and shape completion~\cite{DaiQN17, MittalC0T22}. Besides, several studies have also explored the feasibility of directly generating novel 3D shapes based on implicit representations~\cite{IbingLK21, abs-2209-11163}. However, these approaches are incapable of generating specified 3D shapes that match a given condition, \eg a short text describing the shape characteristics as shown in Figure~\ref{fig:f0}. Text-based visual content synthesis has the advantages of the flexibility and generality~\cite{RameshPGGVRCS21, abs-2204-06125}. Users may generate rich and diverse 3D shapes based on easily obtained natural language descriptors. In addition to generating 3D shapes directly based on text descriptions, manipulating 3D data with text guidance can be further utilized for iterative 3D synthesis and fine-grained 3D editing, which can be beneficial for non-expert users to create 3D visual content.

In the literature, there have been few attempts on the challenging task of text-to-shape generation based on implicit 3D representations~\cite{LiuWQF22, SanghiCLWCFM22, MittalC0T22}. For example, AutoSDF~\cite{MittalC0T22} introduced a vector quantized SDF autoencoder together with an autoregressive generator for shape generation. While encouraging progress has been made, the quality and diversity of generated shapes still requires improvement. The current approaches struggle to generate highly diversified 3D shapes that both guarantee generation quality and conform to the semantics of the input text. Motivated by the success of denoising diffusion models in 2D image~\cite{HoJA20, DhariwalN21, NicholDRSMMSC22} and even explicit 3D point cloud~\cite{LuoH21, ZhouD021, LION2022} generation, we find that DMs achieve high-quality and highly diversified generation while being robust to model training. To this end, we aim to design an implicit 3D representation-based generative diffusion process for text-to-shape synthesis that can achieve better generation flexibility and generalization performance.

In this paper, we propose the \textit{Diffusion-SDF} framework for text-to-shape synthesis based on truncated signed distance fields (TSDFs)\footnote{Strictly speaking, our approach employs a combined explicit-implicit representation in the form of voxelized signed distance fields.}. Considering that 3D shapes share structural similarities at local scales, and the cubic data volume of 3D voxels may lead to slow sampling speed for diffusion models, we propose a two-stage separated generation pipeline. First, we introduce a patch-based SDF autoencoder that map the original signed distance fields into patch-independent local Gaussian latent representations. Second, we introduce the \textit{Voxelized Diffusion} model that captures the intra-patch information along with both patch-to-patch and patch-to-global relations. Specifically, we design a novel \textit{UinU}-Net architecture to replace the standard U-Net~\cite{RonnebergerFB15} for the noise estimator in the reverse process. \textit{UinU}-Net implants a local-focused inner network inside the outer U-Net backbone, which takes into account the patch-independent prior of SDF representations to better reconstruct local patch features from noise. Our work digs deeper into the further potential of diffusion model-based approaches towards text-conditioned 3D shape synthesis based on voxelized TSDFs. Experiments on the largest existing text-shape dataset~\cite{ChenCSCFS18} show that our \textit{Diffusion-SDF} approach achieves promising generation performance on text-to-shape tasks compared to existing state-of-the-art approaches, in both qualitative and quantitative evaluations.

\begin{figure*}
\begin{center}
\includegraphics[width = 1\linewidth]{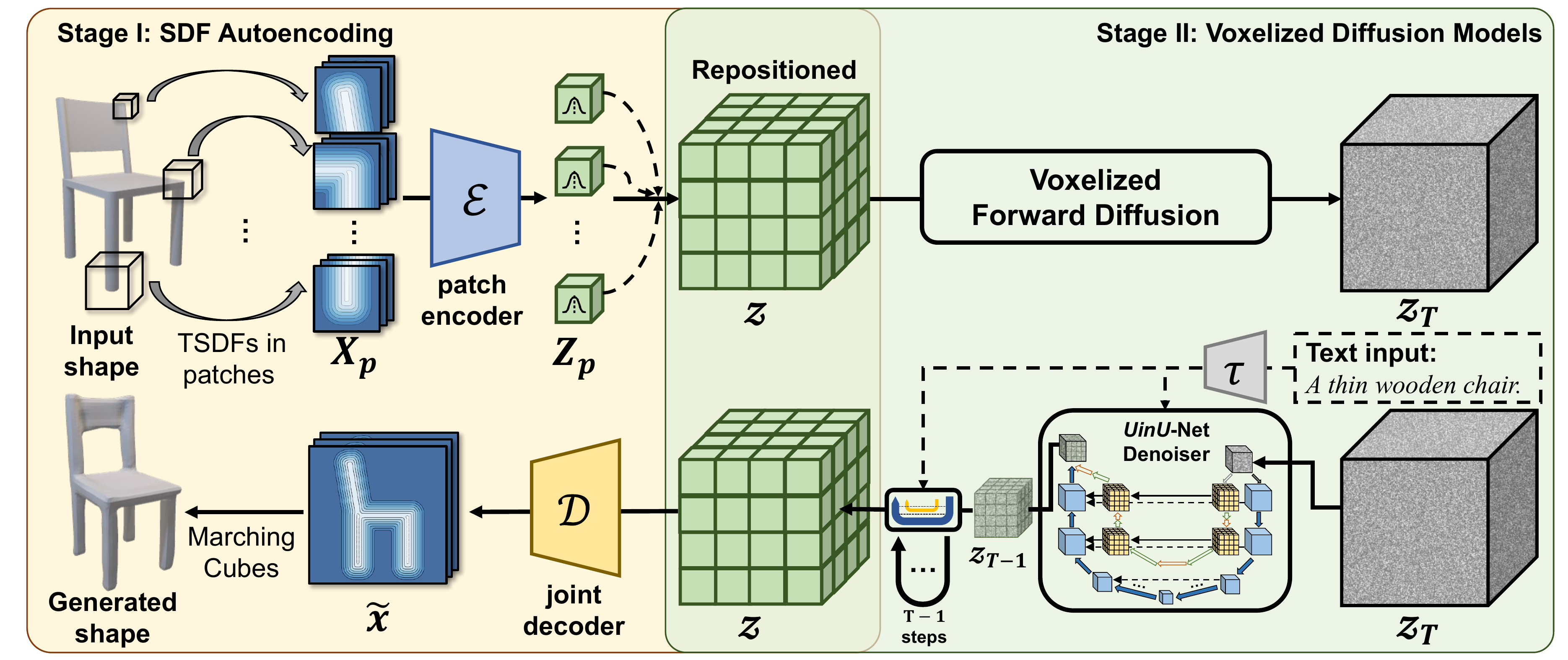}
\end{center}
\vspace{-20pt}
\caption{\textbf{Overview of \textit{Diffusion-SDF}}. We propose a two-stage pipeline to generate novel 3D shapes from texts. First, we train an autoencoder to learn patch-independent normal-distributed representations for voxelized TSDFs. Then, we propose the Voxelized Diffusion framework together with the \textit{UinU}-Net denoiser to generate shape representations. Please check Figure~\ref{fig:snce} for more details of \textit{UinU-Net}.}
\label{fig:pipeline}
\vspace{-5pt}
\end{figure*}

\section{Related Work}
\noindent\textbf{Text-conditioned generative 3D models.}  In the field of generative 3D modeling, a variety of works have focused on synthesizing 3D visual content unconditionally~\cite{0001ZXFT16, AchlioptasDMG18, YangHH0BH19, MeschederONNG19} or conditioned on single/multi-view images~\cite{YanYYGL16, ChoyXGCS16, XuWCMN19, MittalC0T22}. Besides, there has also been a series of works concentrating on the challenging task of text-to-shape generation~\cite{ChenCSCFS18}. Some of them adopted purely explicit 3D data representation-based methods to generate 3D shapes conditioned on input text~\cite{ChenCSCFS18, MichelBLBH22, LION2022}. In contrast, our focus lies on synthesizing 3D shapes based on implicit 3D data representations. So far, there have also been few existing works focusing on the task of implicit text-to-shape generation~\cite{LiuWQF22, SanghiCLWCFM22, MittalC0T22}. All of these approaches have yielded impressive generation results, but there are still some remained issues to be addressed. Sanghi \etal~\cite{SanghiCLWCFM22} proposed a normalizing flow~\cite{RezendeM15}-based approach to generate shape voxels using implicit Occupancy Networks~\cite{MeschederONNG19}, and Liu \etal~\cite{LiuWQF22} introduced a shape-IMLE~\cite{abs-1809-09087} generator using implicit IM-NET~\cite{ChenZ19} decoder, while these approaches neither employ the more flexible implicit SDFs as data representation nor take into account the constraints for local shape structures. More similar to our work, AutoSDF~\cite{MittalC0T22} introduced an autoregressive prior for 3D shape generation based on a discretized SDF autoencoder. The autoregressive model adopts a relatively unnatural way to predict patched 3D tokens in a sequential manner that loses the 3D-specific inductive bias and is also relatively inefficient.

\noindent\textbf{Diffusion Probabilistic Models.} Diffusion Probabilistic Models (DPMs)~\cite{DicksteinW15, HoJA20}, also known as \textit{diffusion models}, have currently arisen as a powerful family of generative models. Compared to previous state-of-the-art generative models, including Generative Adversarial Network (GAN)~\cite{GoodfellowPMXWO20}, Variational Autoencoder (VAE)~\cite{KingmaW13}, and flow-based generative models~\cite{RezendeM15}, the diffusion model demonstrates its superiority in both training stability and generative diversity~\cite{abs-2209-04747}. Diffusion models have achieved promising performance on image~\cite{HoJA20, DhariwalN21, NicholDRSMMSC22, RombachBLEO22} and speech~\cite{ChenZZWNC21, KongPHZC21} synthesis. Especially, fairly impressive results have been achieved with diffusion model-based approaches on the task of text-to-image synthesis~\cite{abs-2204-06125, RombachBLEO22, abs-2205-11487}. In the field of 3D computer vision, several studies have adopted diffusion models for generative 3D modeling~\cite{LuoH21, ZhouD021, LION2022}. PVD~\cite{ZhouD021} employed diffusion models to generate 3D shapes based on point-voxel 3D representation. Luo \etal~\cite{LuoH21} treated points in point clouds as particles in a thermodynamic system with a heat bath. LION~\cite{LION2022} introduced a VAE framework with hierarchical diffusion models in latent space. All these approaches have focused on the diffusion process of explicit 3D data representations for shape generation. On the contrary, our work tries to explore the feasibility of diffusion models on implicit 3D data representations.


\section{Method}

In this section, we introduce the methodology design for \textit{Diffusion-SDF}. In general, we propose a two-stage pipeline as illustrated in Figure~\ref{fig:pipeline}. Detailed information of each stage will be included in the following sections.

\subsection{Autoencoding Signed Distance Fields}

Signed distance fields (SDF) belong to a type of implicit 3D data representation which assigns the scalar signed distance value to the shape surface for each point $p$ in the 3D space $\mathbb{R}^3$. Our objective is to generate the signed distance fields for the target 3D shapes that match the given text conditions. Specifically, the 3D data are represented by truncated signed distance functions (TSDF) in a regularly-spaced voxel grid as~\cite{ChabraLISSLN20, JiangSMHNF20}. Generating voxelized SDFs directly from denoising diffusion models is both cost-expensive and time-consuming. In addition, the local structural information of 3D shapes is not well-emphasized through direct voxel-based generation. To address this, we propose a patch-wise autoencoder to learn the latent representations for voxelized signed distance fields.

Given a 3D shape, we first sample its truncated signed distance field $x$ as a voxel grid of size $D^3$. Before the shape $x$ is encoded, it is first split into $N$ local patches of size $P^3$, producing a sequence of shape patches $X_p=[x_{p1},...,x_{pN}]$, where $N=(D/P)^3$ is the resulting number of patches. Then, the local shape encoder $\mathcal{E}_{loc}$ encodes each shape patches into latent representations $Z_p=[z_{p1},...,z_{pN}]$, where $z_{pn} = \mathcal{E}_{loc}(x_{pn})\in \mathbb{R}^{c}$, and $c$ is the number of latent channels. Here, since each patch is encoded independently, the local structural information can be explored well through the local shape encoder. Meanwhile, the input data scale can also be downsampled by the factor $P$. However, when reconstructing the shape from local patches, it is also necessary to consider the spatial location of each patch in the global shape, and the interrelationship between adjacent patches. To preserve both patch-to-global and patch-to-patch information while decoding the latent embeddings, the patch embeddings are first rearranged into a voxel grid embedding $z$, and then sent into the patch-joint decoder $\mathcal{D}$ that reconstructs the SDF field from the latent patches, giving $\tilde{x}=\mathcal{D}(z)$.

In detail, we adopt a VAE~\cite{KingmaW13, RezendeMW14}-like autoencoder, which encodes each shape patch into a normal distribution. We utilize a combination of both $\mathcal{L}_1$ reconstruction loss together with the KL-regularization loss at the training stage for the SDF autoencoder. The latter one forces a mild KL-penalty towards a standard normal distribution on each learned patch latent. The final SDF representations will be in the form of patch-independent Gaussian distributions.

\subsection{Voxelized Diffusion Models (VDMs)}

\noindent\textbf{Diffusion models (DMs)}~\cite{DicksteinW15, HoJA20} are a class of probabilistic generative models that learn to fit a certain distribution by gradually denoising a Gaussian variable through a fixed Markov Chain of length $T$. Given a data sample $x_0\sim q(x_0)$, DMs describe two different processes in the opposite direction: a \textit{forward process} $q\left(\bm{x}_{0: T}\right)$ that gradually transform a data sample into pure Gaussian noise, and a \textit{reverse process} $p_{\theta}\left(\bm{x}_{0: T}\right)$ that gradually denoise a pure Gaussian sample into real data,
\begin{equation}
\begin{split}
q\left(\bm{x}_{0: T}\right)&=q\left(\bm{x}_{0}\right) \Pi_{t=1}^{T} q\left(\bm{x}_{t} | \bm{x}_{t-1}\right),\\
p_{\theta}\left(\bm{x}_{0: T}\right)&=p\left(\bm{x}_{T}\right) \Pi_{t=1}^{T} p_{\theta}\left(\bm{x}_{t-1} | \bm{x}_{t}\right),
\end{split}
\end{equation}
where $q\left(\bm{x}_{t} | \bm{x}_{t-1}\right)$ and $p_{\theta}\left(\bm{x}_{t-1} | \bm{x}_{t}\right)$ are both Gaussian transition probabilities in the forms of
\begin{equation}
\begin{split}
q\left(\bm{x}_{t} | \bm{x}_{t-1}\right)&=\mathcal{N}\left(\bm{x}_{t};\sqrt{1-\beta_{t}} \bm{x}_{t-1}, \beta_{t} \bm{I}\right),\\
p_{\theta}\left(\bm{x}_{t-1} | \bm{x}_{t}\right)&=\mathcal{N}\left(\bm{x}_{t-1} ;\bm{\mu}_{\theta}\left(\bm{x}_{t}, t\right), \beta_{t} \bm{I}\right).
\end{split}
\end{equation}
Based on~\cite{HoJA20}, the mean variable $\bm{\mu}_{\theta}\left(\bm{x}_{t}, t\right)$ for the reverse transition $p_{\theta}\left(\bm{x}_{t-1}| \bm{x}_{t}\right)$ can be expressed in the form of
\begin{equation}
  \bm{\mu}_{\theta}\left(\bm{x}_{t}, t\right)=\frac{1}{\sqrt{\alpha_{t}}}\left(\bm{x}_{t}-\frac{\beta_{t}}{\sqrt{1-\bar{\alpha}_{t}}} \bm{\epsilon}_{\theta}\left(\bm{x}_{t}, t\right)\right),
\end{equation}
where $\alpha_{t}=1-\beta_{t}$, $\bar{\alpha}_{t}=\Pi_{i=1}^{t} \alpha_{i}$, and $\beta_t$ decreases to 0 gradually as $t$ approaches 0. The evidence lower bound (ELBO) is maximized at the training stage of DMs, eventually leading to the loss function as the form of
\begin{equation}
  \mathcal{L}_{DM} = \mathbb{E}_{\boldsymbol{x}, t, \boldsymbol{\epsilon}\sim\mathcal{N}(0,1)}\left[\left\|\boldsymbol{\epsilon}-\boldsymbol{\epsilon}_{\theta}\left(\bm{x}_t, t\right)\right\|^{2}\right],
\end{equation}
where $\bm{x}_t=\sqrt{\bar{\alpha}_{t}} \boldsymbol{x}_{0}+\sqrt{1-\bar{\alpha}_{t}} \boldsymbol{\epsilon}$, $\boldsymbol{\epsilon}$ is a noise variable, and $t$ is uniformly sampled from $\{1,..., T\}$. The neural network-based score estimator $\boldsymbol{\epsilon}_{\theta}\left(\bm{x}_t, t\right)$ is the core function of denoising diffusion models, which is implemented as a timestep-conditioned denoising autoencoder in~\cite{HoJA20}.

\noindent\textbf{\textit{UinU}-Net Architecture.} In our scenario, we are designing a novel diffusion-based architecture to generate latent SDF representations with regard to the target text conditions. The corresponding TSDF fields can be then reconstructed through the patch-joint decoder which is pre-trained in the first stage. As we mentioned in the above section, the key part of generative diffusion models is to estimate the function approximator $\boldsymbol{\epsilon}_{\theta}$ from the data distribution of the training set. Since our model is designed to generate latent samples, the score estimator can be expressed as $\boldsymbol{\epsilon}_{\theta}\left(z_{t}, t\right)$, as the loss function becomes
\begin{equation}
  \mathcal{L}_{Diffusion-SDF} = \mathbb{E}_{{z}, t, \boldsymbol{\epsilon}\sim\mathcal{N}(0,1)}\left[\left\|\boldsymbol{\epsilon}-\boldsymbol{\epsilon}_{\theta}\left(z_t, t\right)\right\|^{2}\right].
\end{equation}
To incorporate the 3D positional prior of different latent shape patches, we adopt a 3D U-Net~\cite{CicekALBR16}-based autoencoder as the neural function approximator $\boldsymbol{\epsilon}_{\theta}$ that directly learns to denoise the voxel grid embedding $z_t$ that is obtained from the pre-trained local shape encoder $\mathcal{E}_{loc}$. Besides, another crucial prior for our design is the patch-based encoding process, through which all the shape patches are encoded into independent Gaussian distributions. Thus, when generating the local shape embeddings, an important point is to recover the independent distribution for each shape patch as well. To address this, we propose a novel \textit{UinU}-Net architecture for the implementation of the autoencoder-based neural score estimator as shown in Figure~\ref{fig:snce}. The standard 3D U-net~\cite{CicekALBR16} adopts a series of $3\times3\times3$ convolutional layers to construct a downsampling-upsampling network architecture with the information shared through skip connections. This design takes into account the patch-to-patch and patch-to-global features via hierarchical receptive fields. Given that our input data are compressed 3D latent representations, and each patch is encoded independently, we propose to implant another inner network inside the outer U-Net architecture where the embedded resolution is equal to the original latent resolution $D/P$. Specifically, we adopt a $1\times1\times1$ convolution-based ResNet~\cite{HeZRS16} structure to learn independent patch-focused information. We also introduce the spatial Transformer~\cite{VaswaniSPUJGKP17} network that accepts positional embedded patch representations as input tokens, following self-attention layers to capture the relational information between independent local patch embeddings. The inner synthesis paths are skip-connected to the outer ones, ensuring information transmission from inner to outer networks. This architecture is designed to capture the intra-patch information along with both patch-to-patch and patch-to-global relations.
{ 
\begin{figure} \includegraphics[width=1\linewidth] {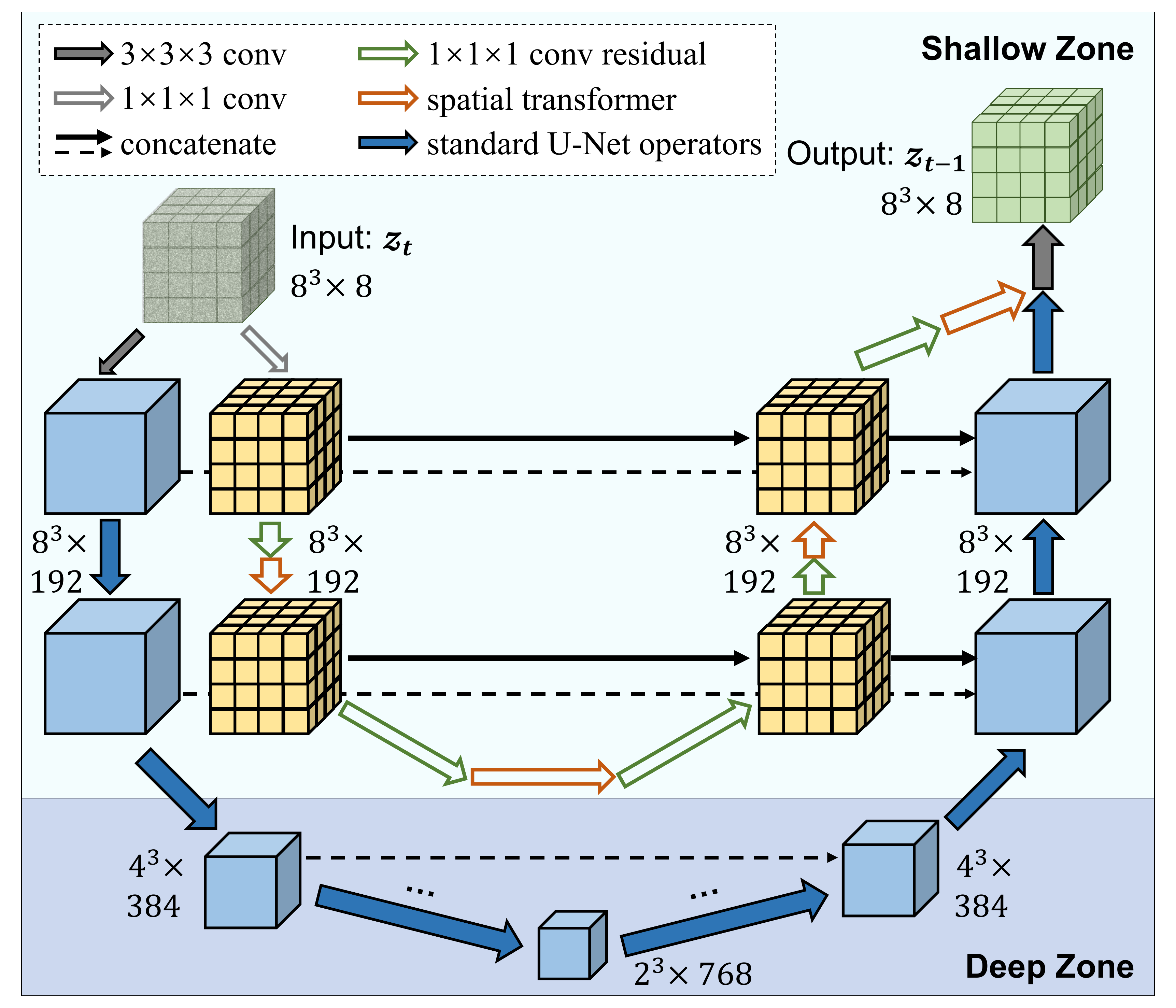} 
\caption{\textbf{Detailed illustration of \textit{UinU-Net}.} To better recover $z$ based on independently distributed patches, we propose to implant an inner network inside the standard U-Net architecture. The inner network is mainly composed of $1\times1\times1$ convolutions, thus it focuses on patch-wise features, while the spatial transformer brings in patch-to-patch information. (Standard U-Net operators include $3\times3\times3$ residual blocks, conditional cross-attention layers, pooling layers and up-scaling layers.)} 
\label{fig:snce} 
\vspace{-5pt} 
\end{figure} }

\noindent\textbf{Text-Guided Shape Generation} So far, we have discussed the generative process without text conditions. To synthesize latent SDF representations based on given conditions, we introduce a conditioning mechanism based on classifier-free guidance diffusion~\cite{ho2021classifierfree}. We adopt a text-conditioned score estimator. In detail, to fit the target conditions into the 3D denoising autoencoder, we utilize the cross-attention mechanism as proposed in~\cite{RombachBLEO22}. The input caption is first encoded as text embeddings through a pre-trained text encoder $\tau$. Thus, the conditioned score can be formed as $\bm{\epsilon}_{\theta}\left(z_{t}|\tau(c)\right)$, where $c$ is the input caption. Meanwhile, we also obtain an unconditioned score $\bm{\epsilon}_{\theta}\left(z_{t}|\emptyset\right)$ with an empty sequence as input caption. Within the guided diffusion sampling process, the output of the model is extrapolated further in the direction of $\bm{\epsilon}_\theta (z_t|\tau(c))$ and away from $\bm{\epsilon}_\theta (z_t|\emptyset)$ as: 
\begin{equation}
\bm{\hat{\epsilon}}_{\theta}\left(z_{t} | \tau(c)\right) = \bm{\epsilon}_{\theta}\left(z_{t} | \emptyset\right)+s \cdot\left(\bm{\epsilon}_{\theta}\left(z_{t} | \tau(c)\right)-\bm{\epsilon}_{\theta}\left(z_{t} | \emptyset\right)\right),
\end{equation}
where $s\geq 1$ refers to the guidance scale. 

\noindent\textbf{Text-Guided Shape Completion} Current shape completion approaches mainly focused on recovering the full shapes from partial input shapes or single view images~\cite{AchlioptasDMG18, YuRWLL021, ZhouD021}. Besides, we have explored the feasibility of text-conditioned shape completion. According to the patch-based design of our SDF encoder, information for the given shape patches can be encoded even if some of the patches are missing. Motivated by the success of diffusion model-based image inpainting~\cite{LugmayrDRYTG22,RombachBLEO22}, we introduce a mask-diffusion strategy to generate the missing shape based on a pre-trained VDM. The core idea is to incorporate known information into each denoising step of the reverse process. For a full shape representation, we transform it into a partial shape representation given the mask $m_z$. For each estimation step $z_{t-1}$, we combine the estimated result through the reverse process $\widetilde{z}_{t-1}$ with the forward sampling result of the unmasked patches $\hat{z}_{t-1}$, 
\begin{equation}
z_{t-1}=(1-m) \odot \widetilde{z}_{t-1}+m \odot 
\hat{z}_{t-1},
\end{equation}
where $\odot$ refers to element-wise product, and
\begin{equation}
\begin{split}
\widetilde{z}_{t-1} &\sim \mathcal{N}\left(\bm{\mu}_{\theta}\left(z_{t}, t, c\right), \beta_{t} \bm{I}\right)\\
\hat{z}_{t-1} &= \sqrt{\bar{\alpha}_{t}} \boldsymbol{z}_{0}+\sqrt{1-\bar{\alpha}_{t}} \boldsymbol{\epsilon}.
\end{split}
\end{equation}
By such a strategy, the unknown region can be recovered based on both the given shape and text conditions.

\noindent\textbf{Text-Guided Shape Manipulation} Visual content editing is one of the most valuable applications for vision-language generation. We propose a diffusion-based text-guided shape manipulation approach. For a given shape representation $z_{init}$, our goal is to transform it into another shape representation $z_{goal}$ based on text $c$. Inspired by image manipulation techniques~\cite{KimKY22a, abs-2210-02249}, we use a cycle-sampling strategy based on a pre-trained VDM. Firstly, we forward-sample the given shape $z_{init}$ for $t_{mid}$ steps, where $0<t_{mid}<T$. This operation will lead to an intermediate output $z_{mid}$
\begin{equation}
z_{mid} = \sqrt{\bar{\alpha}_{t_{mid}}} \boldsymbol{z}_{init}+\sqrt{1-\bar{\alpha}_{t_{mid}}} \boldsymbol{\epsilon}.
\end{equation}
Then, the reverse process will start from $z_{mid}$, and another $t_{mid}$ denoising steps will be conducted conditioned on $c$. Such a design will make the generated shapes correspond to the target text descriptions while maintaining the original shape characteristics.
\begin{figure*}
\begin{center}
\includegraphics[width = 1\linewidth]{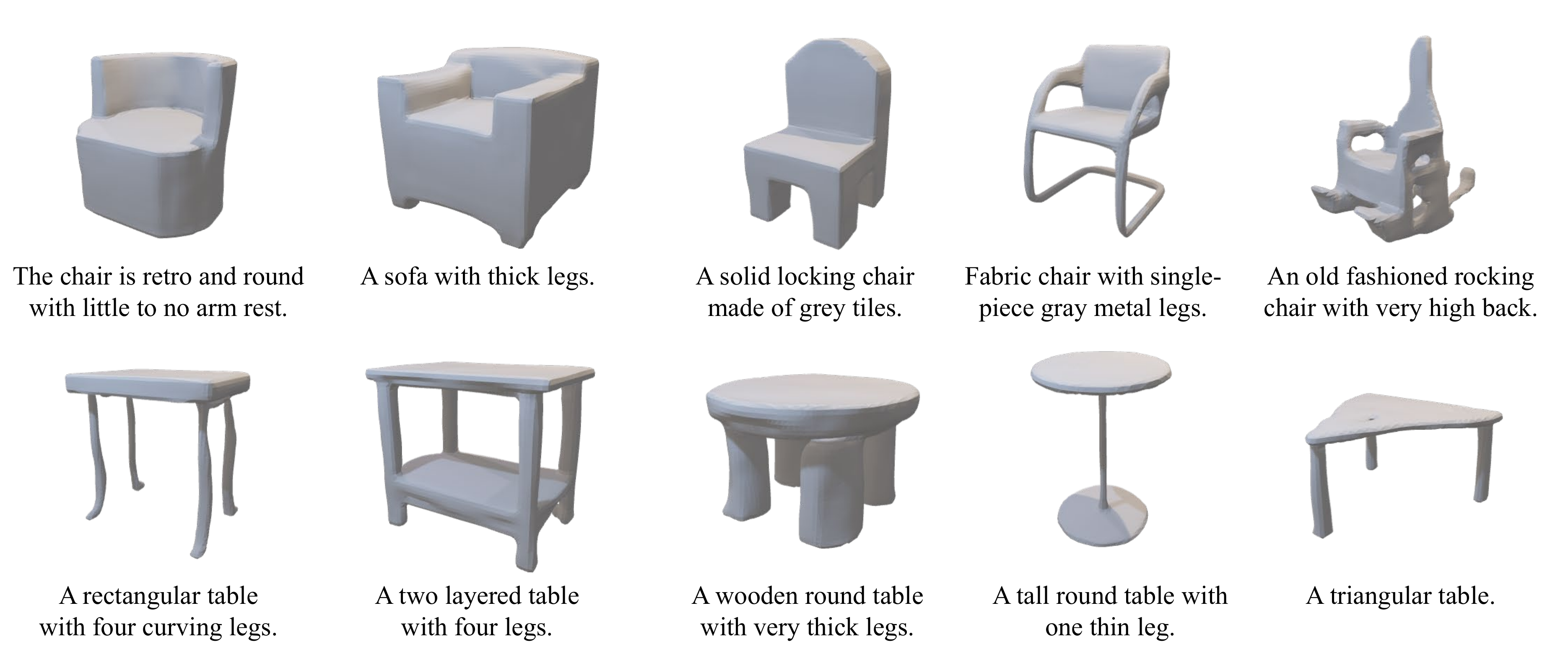}
\end{center}
\vspace{-20pt}
\caption{\textbf{Qualitative results of text-to-shape generation.} Our approach can generate shapes that conform to diverse text descriptions.}
\label{fig:f4}
\vspace{-5pt}
\end{figure*}

\section{Experiments}
We extensively experimented with our proposed \textit{Diffusion-SDF} on various text-conditioned 3D shape synthesis tasks, such as text-to-shape generation, text-conditioned shape completion, and text-guided shape manipulation. The following subsections describe details of experimental settings, results, and analyses.

\noindent\textbf{Dataset.} We mainly trained and evaluated our approach on the current largest text-shape dataset Text2Shape (T2S)~\cite{ChenCSCFS18}. T2S gathered data in the form of shape-text pairs based on two object classes (chairs, tables) in ShapeNet~\cite{ChangFGHHLSSSSX15}. T2S contains about 75K shape-text pairs in total ($\sim$30K for chairs, $\sim$40K for tables), with an average of $\sim$16 words per description. T2S was originally designed to contain both color and shape information within the text descriptions, while we mainly focus on text-shape correspondence.

\noindent\textbf{Implementation details.} For the training stage of the SDF autoencoder, to improve the model's generalization ability, we trained the autoencoder across the 13 categories of ShapeNet~\cite{ChangFGHHLSSSSX15} dataset. For the input data, we sampled the voxelized SDF from original shapes with $64\times64\times64$ grid points, which is also the data size adopted for follow-up operations. For the diffusion stage, to facilitate the generative sampling speed, we adopt the DDIM~\cite{SongME21} sampler to reduce the original DDPM sampling steps from $T=1000$ to $50$. For the conditioning mechanism, we adopt a freeze CLIP~\cite{RadfordKHRGASAM21} text embedder as text encoder $\tau$.

\subsection{Text-Conditioned Shape Generation}
The most immediate application of our approach involves generating novel shapes that are conditioned on textual descriptions. To showcase the efficacy of our approach, we conduct a comparative evaluation of our generation results against two state-of-the-art supervised text-to-shape synthesis methods that are based on implicit 3D representations~\cite{LiuWQF22, MittalC0T22}. Besides, there are some other works regarding text-to-shape generation, which have different settings compared to our approach. For instance,~\cite{SanghiCLWCFM22} proposed a zero-shot shape generation approach conditioned on categorical texts, and~\cite{MichelBLBH22, abs-2209-11163} proposed to generate textured shapes based on provided/pre-generated meshes. As a result, these prior works are not directly compared against our proposed approach.

\noindent\textbf{Quantitative comparison.} To compare the generation performance of our approach to the previous methods quantitatively, we propose to use several metrics to evaluate the generated results:
\begin{enumerate}[-]
\item \textbf{IoU} (Intersection over Union) measures the occupancy similarity between the generated shape to the ground truth. To compute the IoU score, we downsample all the generated SDFs to occupancy voxel grids of size $32^3$. IoU metric is used to measure the conformity of the generated samples with the ground truth shapes.
\item \textbf{Acc} (Classification Accuracy) introduces a voxel-based classifier, which is pre-trained to classify the 13 categories in ShapeNet~\cite{ChangFGHHLSSSSX15} according to~\cite{SanghiCLWCFM22}. Acc metric is used to measure the semantic authenticity of the generated samples.
\item \textbf{CLIP-S} (CLIP Similarity Score) introduces the pre-trained vision-language model CLIP~\cite{RadfordKHRGASAM21} for further evaluation. The CLIP can be used to measure the correspondence between given images and texts. We render 5 view images for each generated shape, and compute the cosine similarity score between rendered images and the given text. The highest score is reported for each text query. CLIP-S metric is used to measure the semantic conformance between the generated samples and input texts.
\item \textbf{TMD} (Total Mutual Difference). For each given text description, we generate $k=10$ different samples. Then, we compute the average IoU score for each generated shape to other $k-1$ shapes. The average IoU score for all text queries is reported. TMD assesses the generation diversity for each given text query.
\end{enumerate}
To keep consistency with previous approaches, we limit the comparisons to the chair category in Text2Shape~\cite{ChenCSCFS18} dataset. Since the officially released AutoSDF~\cite{MittalC0T22} model was trained on another dataset~\cite{AchlioptasGGFH19}, we re-trained the model following the original settings on the training set of Text2Shape for the fair comparison. The results are shown in Table~\ref{table:1}. From the results, it can be seen that our approach is able to achieve relatively high generation quality with high IoU and Acc scores, and robust text-shape conformance with good CLIP-S performance, while achieving much better generation diversity with a much lower TMD performance.

\begin{table}[]
\small
\begin{center}
\setlength{\tabcolsep}{5pt}
\caption{Quantitative comparisons of text-to-shape generation.}
\vspace{-5pt}
\begin{tabular}{lcccc}
\toprule
\textbf{Methods}  & \textbf{IoU↑} & \textbf{Acc↑} & \textbf{CLIP-S↑} & \textbf{TMD↓} \\ \hline
Liu \etal~\cite{LiuWQF22} &  0.160             &      34.79        &     29.94            &    0.891          \\
AutoSDF~\cite{MittalC0T22}           & 0.187        & 83.88        & 29.10           & 0.581        \\
Diffusion-SDF (Ours) & \textbf{0.194}        & \textbf{88.56}        & \textbf{30.88}           & \textbf{0.169}      \\ 
\bottomrule
\end{tabular}\label{table:1}
\vspace{-15pt}
\end{center}
\end{table}

\begin{figure*}
\begin{center}
\includegraphics[width = 1\linewidth]{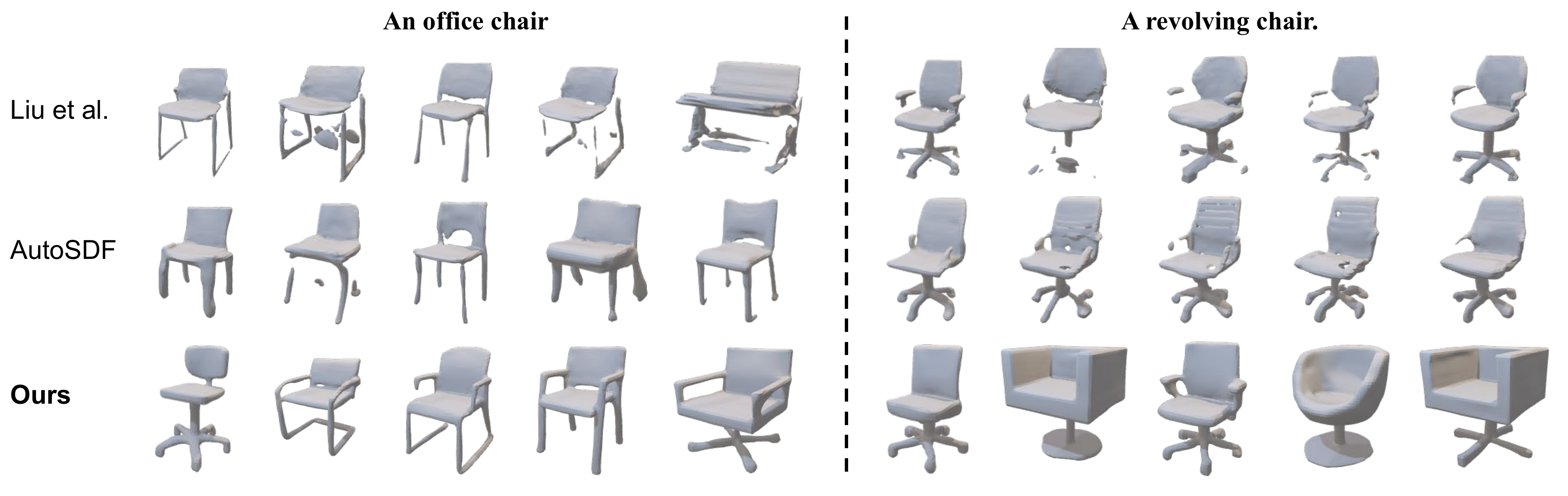}
\end{center}
\vspace{-20pt}
\caption{\textbf{Qualitative comparison of diversified text-to-shape generation.} Our approach is capable of generating highly-diversified shapes that conform to the same text query, while also keeping the satisfactory quality of the generated shapes.}
\label{fig:f5}
\vspace{-5pt}
\end{figure*}

\noindent\textbf{Qualitative results.} The quantitative experiment shows the generation performance of our approach on the limited given subset from Text2Shape dataset. Due to the generalization of natural language, our approach is capable of generating different forms of shapes based on various text descriptions. Figure~\ref{fig:f4} shows the generated results of \textit{Diffusion-SDF} from different text inputs. The results show that our method can generate eligible results from distinguished conditions. Besides, our method is capable of synthesizing various shape outputs from the same input text description. Figure~\ref{fig:f5} compares our approach with~\cite{MittalC0T22, LiuWQF22} based on same text queries. The results have shown that our method has a great advantage in the diversity of generated samples, while the previous approaches are unable to generate highly-diversified shapes. 

\begin{figure*}
\begin{center}
\includegraphics[width = 1\linewidth]{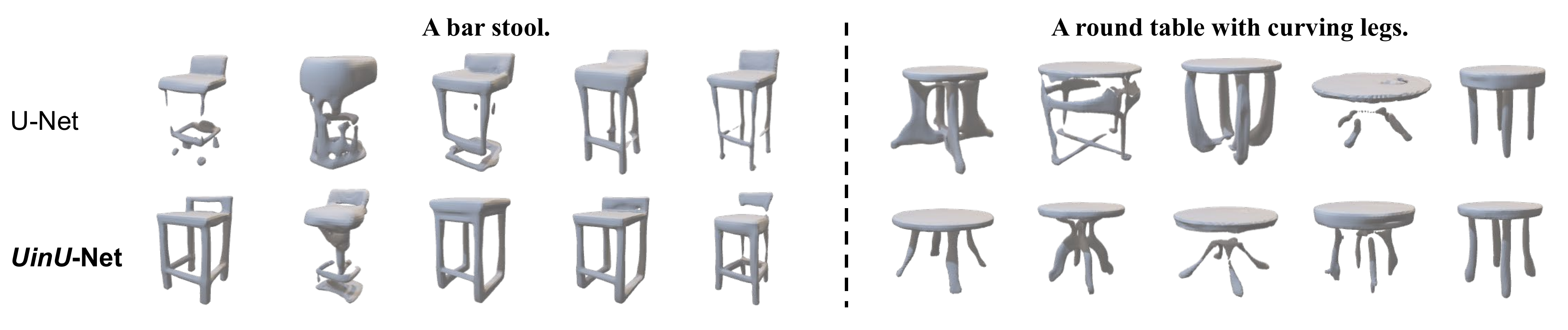}
\end{center}
\vspace{-20pt}
\caption{\textbf{Qualitative comparison of U-Net and \textit{UinU}-Net.} Since \textit{UinU}-Net-based architecture contains a local-focused inner network, it can better recover the independent patch representations, thereby improving the generation quality compared to conventional U-Net.}
\label{fig:f6}
\vspace{-5pt}
\end{figure*}

\subsection{Ablation Studies}
We have conducted extensive ablation studies to demonstrate the effectiveness of our special architecture design for \textit{Diffusion-SDF}.

\noindent\textbf{Effectiveness of the \textit{UinU}-Net architecture.} In order to validate the benefit of our proposed modifications to the conventional U-Net architecture, we additionally compare our proposed \textit{UinU}-Net architecture to the simpler U-Net architecture used in~\cite{RombachBLEO22}. The quantitative results for text-to-shape generation are shown in Table~\ref{table:2}. Qualitative comparisons are shown in Figure~\ref{fig:f6}. From the results, it can be found that our proposed \textit{UinU}-Net architecture for Voxelized Diffusion models generally improves the quality of generated shapes. With the \textit{UinU}-Net architecture, the noise for each step is estimated both patch-mutually and patch-independently. Since the shape representations are repositioned based on separate patch embeddings extracted by the SDF encoders, the key to recovering the authentic shape SDFs from the joint decoder is to ensure that the generated SDF representations are also distributed independently in patches. From the qualitative results, it can be inferred that the inner network is conducive to recovering the independently distributed patch representations.

\noindent\textbf{Inner network architecture.} As for the \textit{UinU}-Net architecture, we have conducted ablation experiments on several variants for the inner network to validate the effectiveness of the final design. The results have shown in Table~\ref{table:3}. According to the results, it can be inferred that the inner outer concatenation mechanism and spatial attention layer can generally improve the generation quality. Without the inner-outer concatenation mechanism, the accuracy score and CLIP-S have significant decreases, indicating that the information transmission mechanism has affected the semantic authenticity and conformance of the generated shapes. Accuracy score has dropped without the spatial attention module, indicating that the introduction of patch-to-patch information helps preserve the full shape semantics while recovering the patch-wise representations.

\begin{table}[]
\small
\begin{center}
\setlength{\tabcolsep}{5pt}
\caption{Evaluating effectiveness of the \textit{UinU}-Net architecture.}
\vspace{-5pt}
\begin{tabular}{lcccc}
\toprule
  & \textbf{IoU↑} & \textbf{Acc↑} & \textbf{CLIP-S↑} & \textbf{TMD↓} \\ \hline
U-Net~\cite{RombachBLEO22}           & 0.187        & 80.98        & 29.35           & 0.171        \\
\textit{UinU}-Net (Ours) &  \textbf{0.194}        & \textbf{88.56}        & \textbf{30.88}           & \textbf{0.169}      \\ 
\bottomrule
\end{tabular}\label{table:2}
\vspace{-15pt}
\end{center}
\end{table}

\begin{table}[]
\small
\begin{center}
\setlength{\tabcolsep}{5pt}
\caption{Evaluating effectiveness of the inner network design.}
\vspace{-5pt}
\begin{tabular}{lcccc}
\toprule
  & \textbf{IoU↑} & \textbf{Acc↑} & \textbf{CLIP-S↑} & \textbf{TMD↓} \\ \hline
\textit{UinU}-Net               &  0.194        & \textbf{88.56}        & \textbf{30.88}           & 0.169      \\ 
- \textit{w/o} in-out concat     & \textbf{0.198}              & 83.91             &  28.68               &  0.181          \\
- \textit{w/o} spatial attention &   0.194            &     85.56         &    30.82             &      \textbf{0.161}       
\\ 
\bottomrule
\end{tabular}\label{table:3}
\vspace{-15pt}
\end{center}
\end{table}

\begin{figure*}
\begin{center}
\includegraphics[width = 1\linewidth]{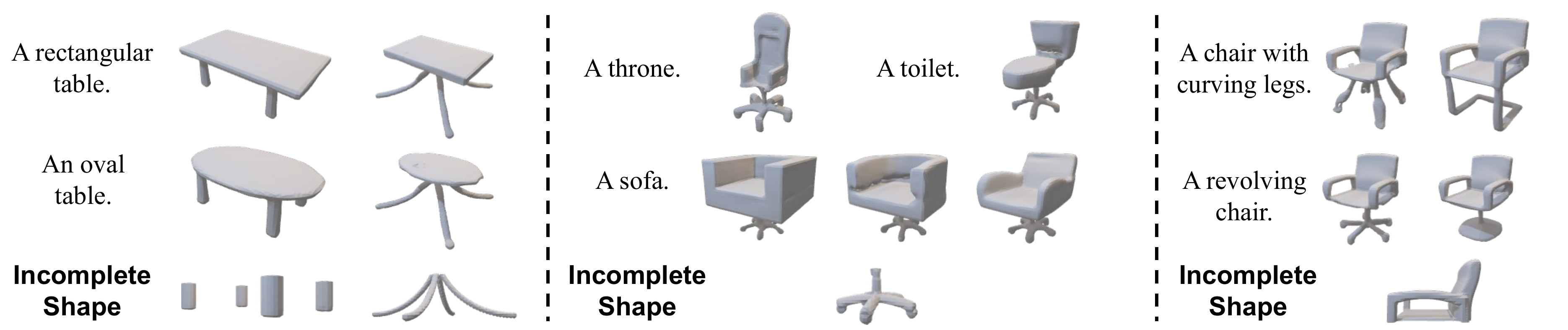}
\end{center}
\vspace{-15pt}
\caption{\textbf{Qualitative result of text-guide shape completion.} Given a known partial shape, our approach can generate the missing part based on text conditions.}
\label{fig:f7}
\vspace{-5pt}
\end{figure*}

\begin{figure*}
\begin{center}
\includegraphics[width = 1\linewidth]{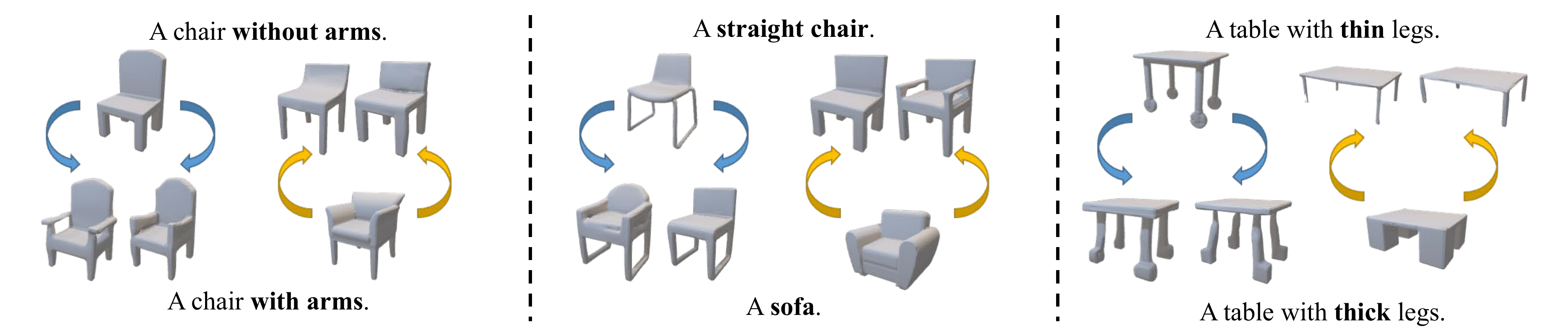}
\end{center}
\vspace{-15pt}
\caption{\textbf{Qualitative result of text-guide shape manipulation.} Given a known shape, our approach is able to manipulate the given shape into the target shape based on the text descriptions.}
\label{fig:f8}
\vspace{-5pt}
\end{figure*}

\subsection{Text-Guided Shape Completion}
In this section, we validate the effectiveness of our approach on the text-guided shape completion task. Specifically, given a partial input shape, our goal is to generate the missing part conditioned on the input text descriptions. For instance, we can generate the body of a chair based on the given chair legs or generate the missing chair legs based on the chair body. The qualitative results of our approach are displayed in Figure~\ref{fig:f7}. The results show that our approach is capable of generating the missing part of a shape while being well-blended with the given shape, as evident in the generated examples such as \textit{revolving throne}, \textit{revolving toilet}, and \textit{revolving sofa}. Moreover, our completion performance is not limited by the shape geometry in the training set. The core limitation of our approach is the cut region of patch grids, which can be addressed by increasing resolution and decreasing patch size.

\subsection{Text-Guided Shape Manipulation}
We demonstrate our approach's effectiveness for text-guided shape manipulation. Our approach can modify a given shape to match an instructive text description. Figure~\ref{fig:f8} gives several qualitative illustrations for our proposed approach. From the results, it can be summarized that our approach can handle the shape manipulation of both local shape structures (\eg the existence/style of arms or legs) or the overall shape characteristics (\eg sofa or straight chair). In addition, one key issue for our designed approach is the choice for $t_{mid}$. Since the shape characteristics usually form in the early stage of the reverse process, thus we set $t_{mid}$ to relatively larger step numbers (600$\sim$800 step before DDIM's down-sampling). In experiments, we observe that excessively small or large values for $t_{mid}$ result in unchanged or destructed original shape features, respectively.

\section{Limitations and Conclusion}

\textbf{Limitations. }This paper demonstrates the superiority of our approach on the Text2Shape~\cite{ChenCSCFS18} dataset, which is limited to only two categories from ShapeNet, restricting our approach's generalization to other categories. Moreover, the lack of current shape-text datasets prevents us from validating our approach on additional benchmarks. To address this, we propose introducing more datasets and exploring zero-shot text-to-shape generation by leveraging knowledge from 2D vision-language models. While some studies have attempted to tackle this challenge~\cite{SanghiCLWCFM22}, achieving flexible text-to-shape generation remains a significant obstacle. Our future work will seek a possible solution to this problem.

\textbf{Conclusion. }Our paper presents a framework for text-to-shape synthesis called \textit{Diffusion-SDF}, which utilizes a diffusion model to generate voxelized SDFs conditioned on texts. The approach comprises a two-stage pipeline: a patch-wise autoencoder for generating Gaussian SDF representations, followed by a Voxelized Diffusion model with \textit{UinU}-Net denoisers for generating patch-independent SDF representations. We evaluate our approach on various text-to-shape synthesis tasks, and the results demonstrate that it can generate highly diverse and high-quality 3D shapes.

\noindent\textbf{Acknowledgments    } This work was supported in part by the National Key Research and Development Program of China under Grant 2017YFA0700802, in part by the National Natural Science Foundation of China under Grant 62206147 and Grant 62125603, and in part by a grant from the Beijing Academy of Artificial Intelligence (BAAI).

\renewcommand\thesection{\Alph{section}}
\setcounter{section}{0}

\section{Qualitative Illustrations}
\begin{figure*}
\begin{center}
\includegraphics[width = 1\linewidth]{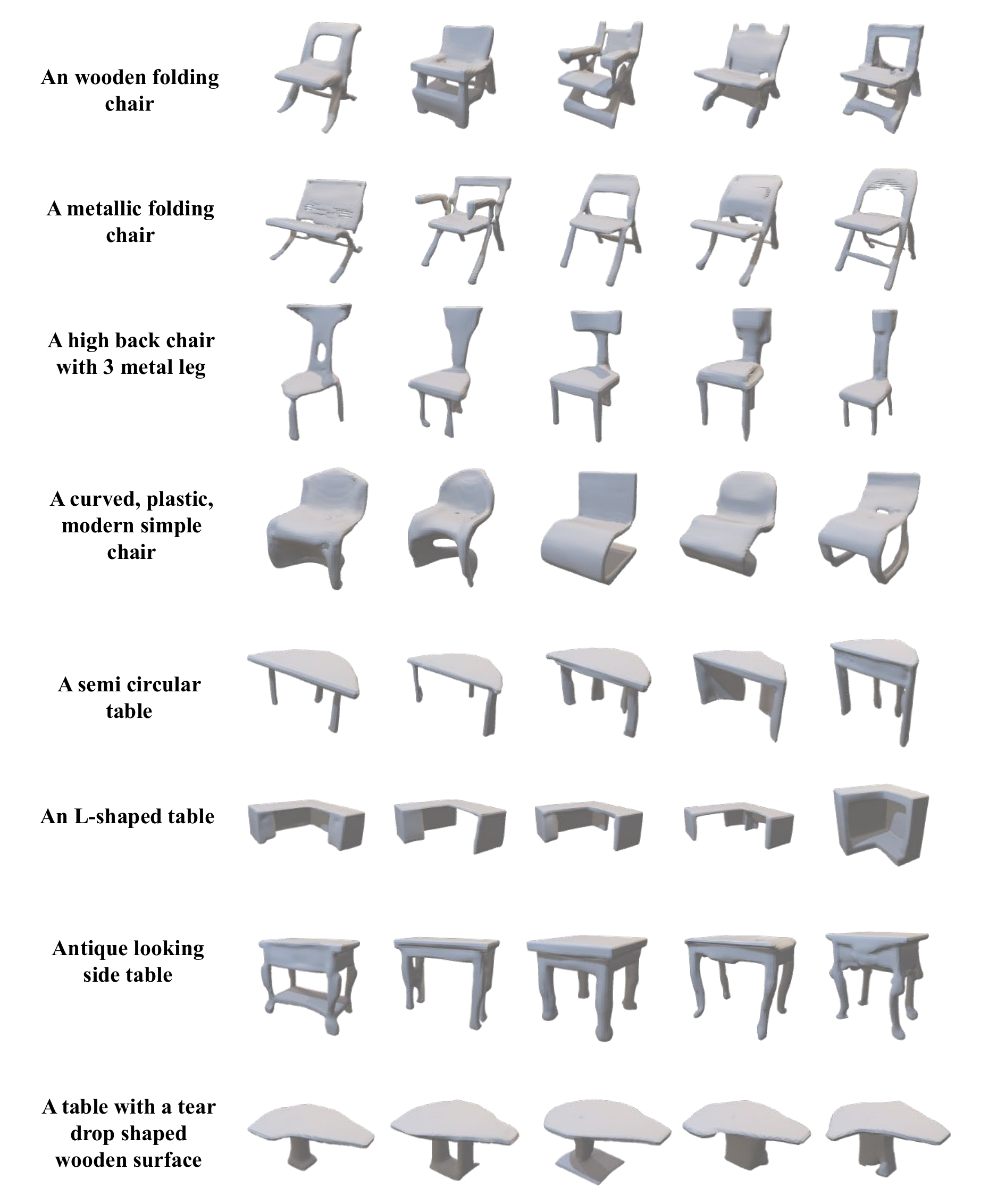}
\end{center}
\vspace{-20pt}
\caption{More illustrations on diversified text-to-shape generation.}
\label{fig:1}
\vspace{-5pt}
\end{figure*}

\begin{figure*}
\begin{center}
\includegraphics[width = 1\linewidth]{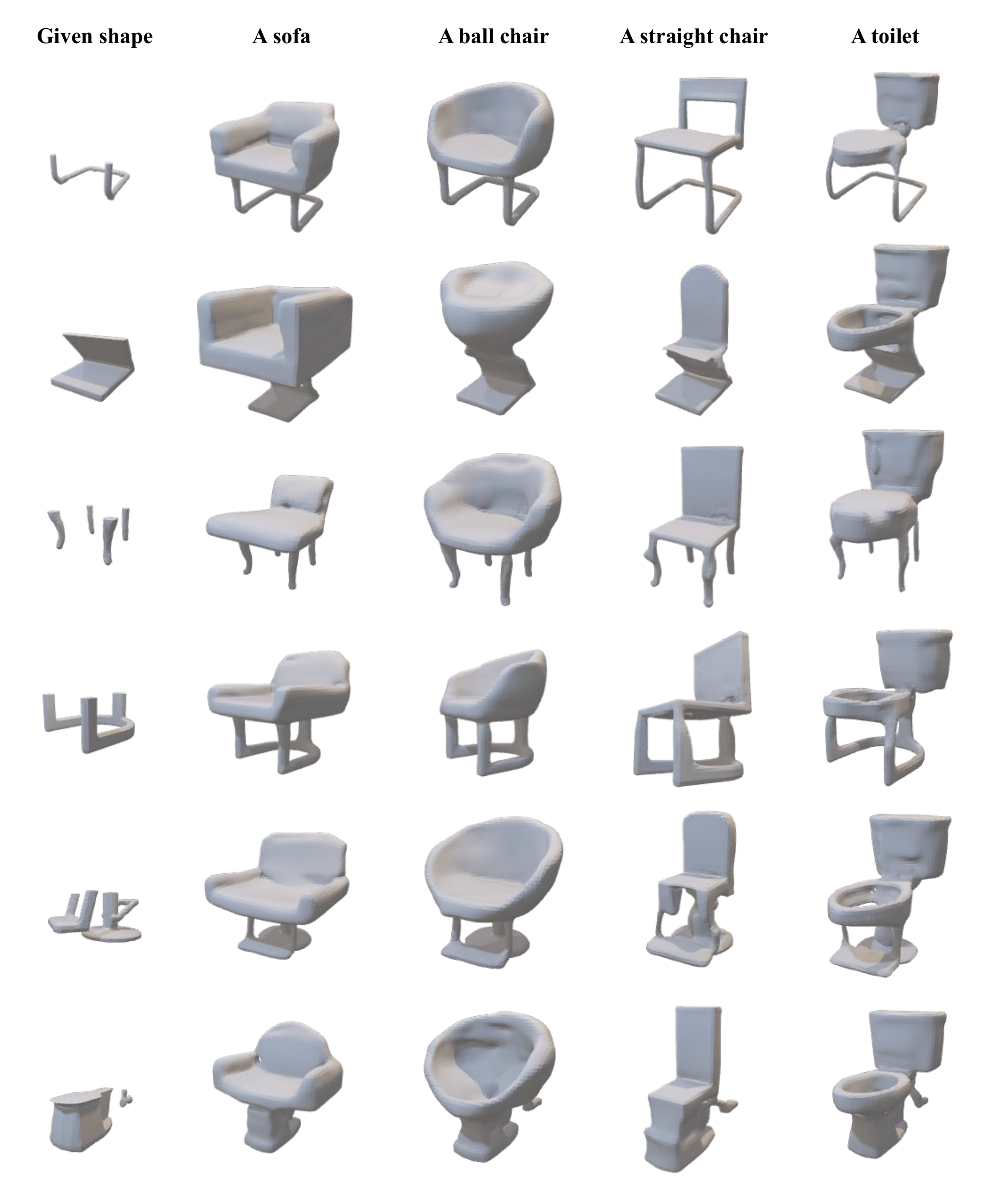}
\end{center}
\vspace{-20pt}
\caption{More illustrations on text-conditioned shape completion.}
\label{fig:2}
\vspace{-5pt}
\end{figure*}

\begin{figure*}
\begin{center}
\includegraphics[width = 1\linewidth]{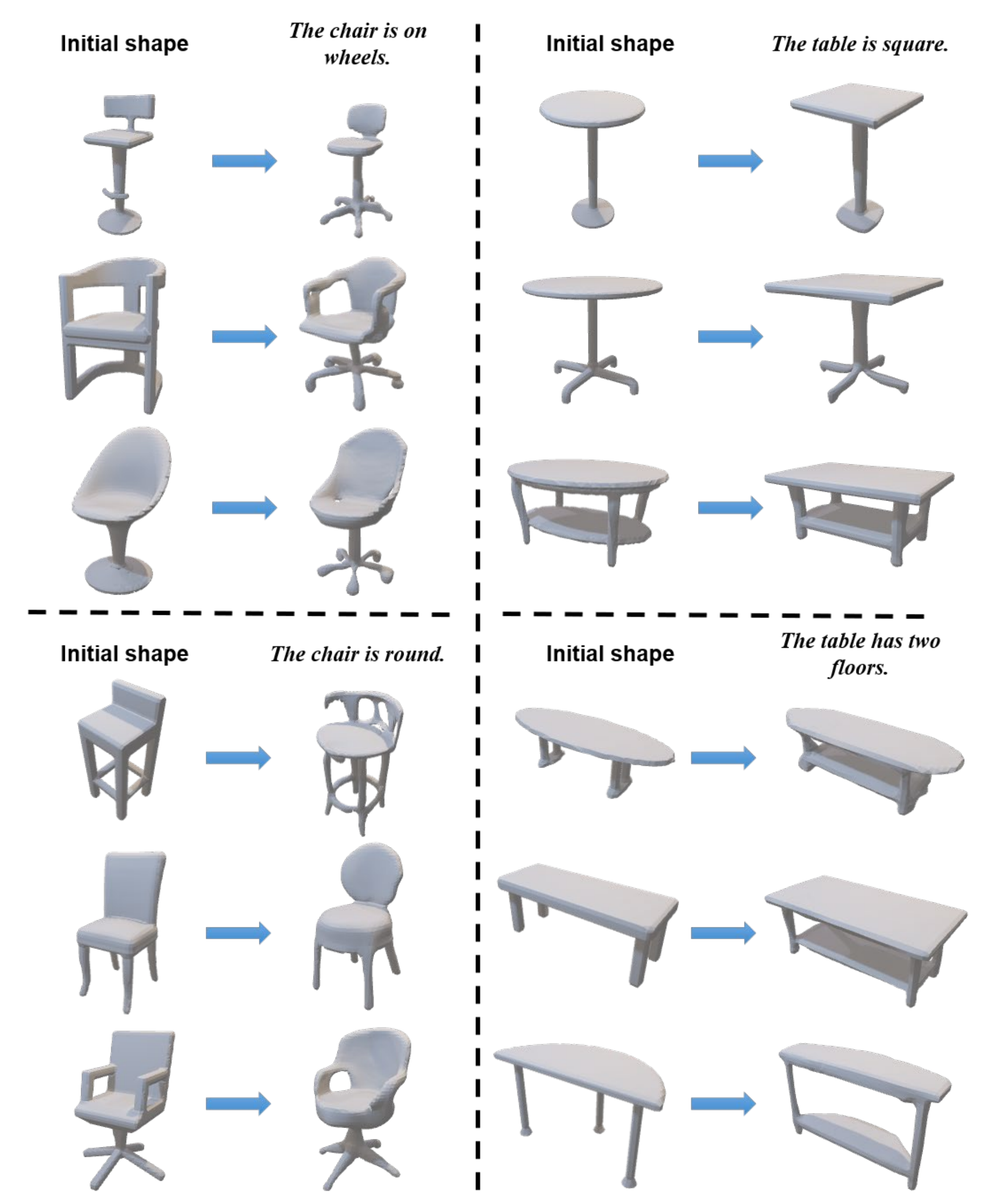}
\end{center}
\vspace{-20pt}
\caption{More illustrations on text-conditioned shape manipulation.}
\label{fig:3}
\vspace{-5pt}
\end{figure*}

In this section, we provide more qualitative illustrations for the generation results based on our proposed \textit{Diffusion-SDF} approach. Figure~\ref{fig:1} shows some extra diversified generated samples for text-conditioned shape generation. Figure~\ref{fig:2} displays further text-conditioned shape completion results based on different input shapes. Figure~\ref{fig:3} illustrates several text-conditioned shape manipulation results from diverse initial shapes. The extra qualitative illustrations are displayed at the end of this paper.

{\small
\bibliographystyle{ieee_fullname}
\bibliography{egbib}
}

\end{document}